\documentclass{article}
\usepackage{spconf,amsmath,graphicx}
\usepackage{amsmath}
\usepackage{multirow}
\usepackage{comment}
\usepackage{verbatim}
\usepackage{xcolor}
\usepackage{comment}
\usepackage{caption}
\usepackage{amsfonts}
\usepackage{subfigure}
\usepackage{enumitem}

\title{A Lightweight Feature Fusion Architecture for Resource-constrained Crowd counting}
\name{
  \begin{tabular}{c}
    Yashwardhan Chaudhuri\textsuperscript{1*}, Ankit Kumar\textsuperscript{2*}, Orchid Chetia Phukan\textsuperscript{1}, 
    Arun Balaji Buduru\textsuperscript{1}
  \end{tabular}\thanks{*Authors contributed equally as first authors}
}

\address{
  \begin{tabular}{c}
    \textsuperscript{1}IIIT Delhi, India \\
     \textsuperscript{2}IIT Bombay, India
  \end{tabular}
}
\begin{document}
%
\maketitle

\begin{abstract}
Crowd counting finds direct applications in real-world situations, making computational efficiency and performance crucial.  However, most of the previous methods rely on a heavy backbone and a complex downstream architecture that restricts the deployment. To address this challenge and enhance the versatility of crowd-counting models, we introduce two lightweight models. These models maintain the same downstream architecture while incorporating two distinct backbones: MobileNet and MobileViT. We leverage Adjacent Feature Fusion to extract diverse scale features from a Pre-Trained Model (PTM) and subsequently combine these features seamlessly. This approach empowers our models to achieve improved performance while maintaining a compact and efficient design. With the comparison of our proposed models with previously available state-of-the-art (SOTA) methods on ShanghaiTech-A  ShanghaiTech-B and UCF-CC-50 dataset, it achieves comparable results while being the most computationally efficient model. Finally, we present a comparative study, an extensive ablation study, along with pruning to show the effectiveness of our models.
\end{abstract}
\section{Introduction}
\begin{figure*}[ht]
    \centering
    \subfigure[ASFNet-S]{%
         \includegraphics[width=0.45\textwidth]{ 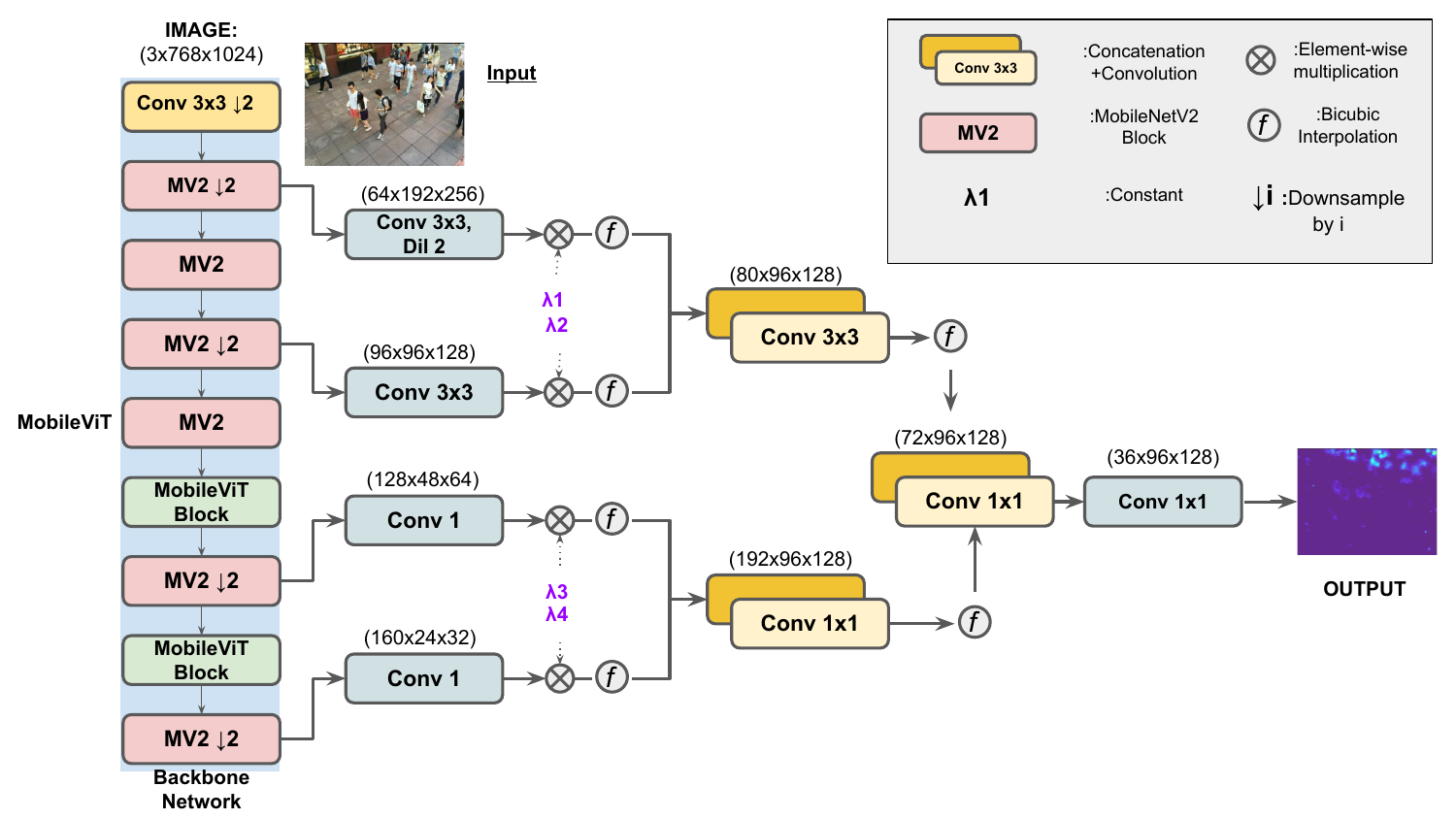}
        \label{fig:subfiga}
    }
\hfill
    \subfigure[ASFNet-B]{%
       \includegraphics[width=0.45\textwidth]{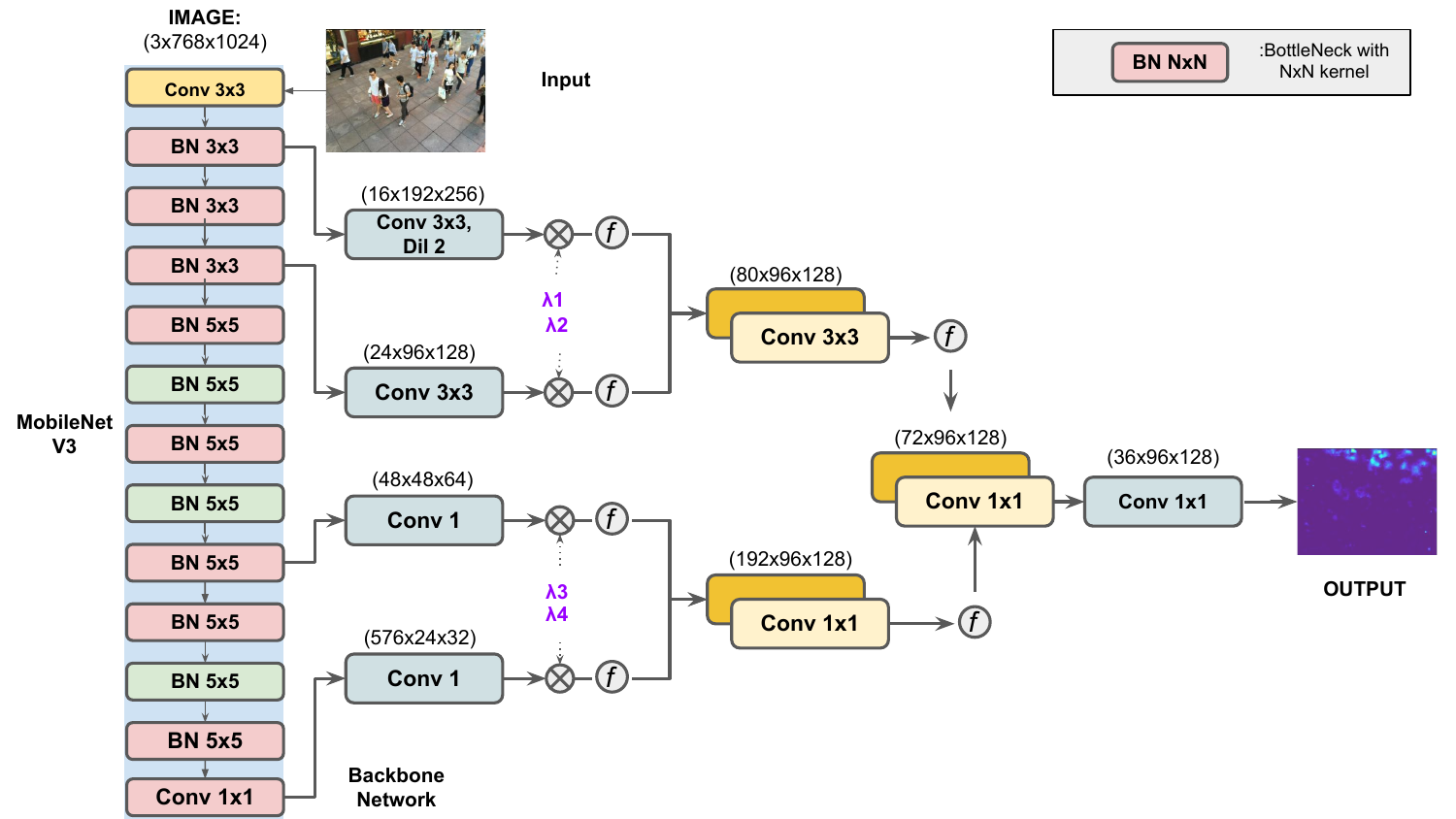}
        \label{fig:subfigb}
    }
    \caption{\small The overall architectural depiction of ASFNet-B and ASFNet-S illustrates ASFNet-S leverages a MobileViT backbone, while ASFNet-B employs a MobileNet backbone, common downstream shared by both networks, enabling multi-scale feature extraction.  }
    \label{figure_1}
\end{figure*}

Crowd counting refers to the estimation of the number of people in a crowd scene from either an image or a video, which found numerous applications that hold significant societal relevance, including urban planning, area footfall estimation, and crowd management. Crowd counting is a challenging problem because of the diverse behavior of crowds, heavy occlusions, and complex backgrounds. Recently, CNN-based networks have made remarkable progress, having a heavy architecture that demands substantial resources and computation costs, which restricts their deployment scopes and causes poor scalability. 
Many methods focus on dealing with this problem propose lightweight architectures, utilize pruning \cite{anwar2017structured} \cite{peng2019collaborative} and  Quantization\cite{lin2016fixed} \cite{yang2019quantization}, Although effective, these following methods suffer from meticulous hyperparameter tuning and hardware limitations. Moreover, the following methods can also significantly affect model generalization due to a regularization effect.
Other methods \cite{liu2020efficient} \cite{liu2020metadistiller} utilize knowledge distillation-based method that makes use of supervision from a large model (teacher) to train a smaller model (student). While this technique enhances the performance of the student model, it also requires additional training time and careful selection of teacher models for better learning. These methods push heavy models to become lightweight, but the inherent computational resource requirement and model size remain massive from the perspective of existing lightweight works. To address these challenges, We propose Adjoining Semantics Fusion Network(ASFNet), a lightweight crowd-counting network generating density maps by progressively fusing adjacent feature sets at each level, as shown in figure \ref{figure_1}. ASFNet comes in two variations: ASFNet-B and ASFNet-S.  Both of these architectures share a common downstream network but are characterized by their distinct backbones. It involves a small number of parameters and achieves satisfying performance on benchmark datasets. The key contributions are as follows:

\begin{itemize}[noitemsep]
  \item We propose ASFNet-S and ASFNet-B, two varieties of lightweight architectures with different backbones and common downstream networks for resource-efficient crowd-counting. We utilize the adjacent feature fusion technique, which merges different scale features for easier feature integration and improved learnability.
  
  \item Evaluation of our models on benchmark datasets such as ShanghaiTech-A, ShanghaiTech-B and UCF-CC-50, shows comparable performance with SOTA works across various metrics with significantly less parameters, FLOPs and inference time.
   
 \item  Ablation study to find the efficacy of our model; through extensive experiments, ablation study, and pruning analysis, we confirm the viability of our model architecture. 

\end{itemize}
\vspace{-3.3mm}

\section{Related work}
Most recent crowd-counting methods employ CNN to predict density maps from crowd images. MCNN \cite{zhang2016single} employs a multi-column architecture that effectively captures scale variation. CSRNet \cite{li2018csrnet} CANNET\cite{liu2019context} use single-column with dilated convolution layer. In \cite{cheng2022rethinking}, the convolution filter is replaced by locally connected Gaussian kernels called GauNet and proposes a low-rank approximation for the translation invariance problem. MSSRM\cite{xie2023super} propose a  Multi-Scale Super-Resolution Module that guides the network to estimate the lost details and enhances the detailed information. These methods perform well, but they are large in size and computationally expensive. Some lightweight architecture \cite{shi2020real} \cite{wang2022eccnas} \cite{9190692} use feature fusion and quantization, which are hardware-dependent architecture.\cite{liu2020efficient} purpose Structured Knowledge Transfer (SKT) framework that maximizes the utilization of structured knowledge from a proficiently but depends on a trained teacher network to produce a streamlined student network.Lw Count\cite{liu2022lw}  SANet\cite{cao2018scale} propose an encoding-decoding-based lightweight crowd-counting network. Furthermore, in our work, we explore,

\vspace{-0.6cm}
\section{Methodology}
\subsection{Proposed Architecture}
We propose two different architectures, ASFNet-S and ASFN-et-B, as shown in Figure \ref{figure_1}, having the same downstream network with different backbones, i.e. MobileViT\footnote{https://huggingface.co/apple/mobilevit-small}\cite{wadekar2022mobilevitv3} for ASFNet-B and MobileNetV3 \footnote{https://download.pytorch.org/models/mobilenet\_v3\_small-047dcff4.pth}\cite{howard2017mobilenets} for ASFNet-S. Each has two key components: a pre-trained backbone for feature extraction and a downstream network with adjacent feature fusion responsible for density map generation. The pretrained backbone is typically a deep neural network previously trained on a large dataset imagenet, serving to extract multiscale features from input data. The downstream network follows the backbone and further processes these features, with adjacent feature fusion indicating that feature fusion happens in a nearby and sequential manner within the network. This fusion combines features from different levels of abstraction, allowing the network to benefit from low-level and high-level information for density map generation.  
\vspace{-1.3mm}
\subsection{Feature Extraction}MobileNet and MobileVit are well-suited backbone networks for lightweight crowd-counting tasks primarily because of their efficiency and speed, making them an excellent choice for resource-constrained and real-time applications. MobileNetV3 is an efficient convolutional neural network designed for fast deep learning on mobile devices. It achieves efficiency through depthwise separable convolutions and maintains accuracy with inverted residual blocks.
The MobileViT block combines standard convolutions and transformers to capture local and global representations. It Replaces local processing with a stack of transformer layers gains convolution-like properties while learning global context efficiently. 
We utilize the four blocks of the backbone network to capture diverse feature information at different levels of semantic information, with initial layers having low-level semantic information and the later layers having high-level semantic information. The output of the feature extraction block is passed sequentially to the downstream network, which uses the adjacent feature fusion technique for density map generation.

\begin{figure*}[ht!]
  \centering
  \includegraphics[width=\textwidth, height=0.20\textwidth]{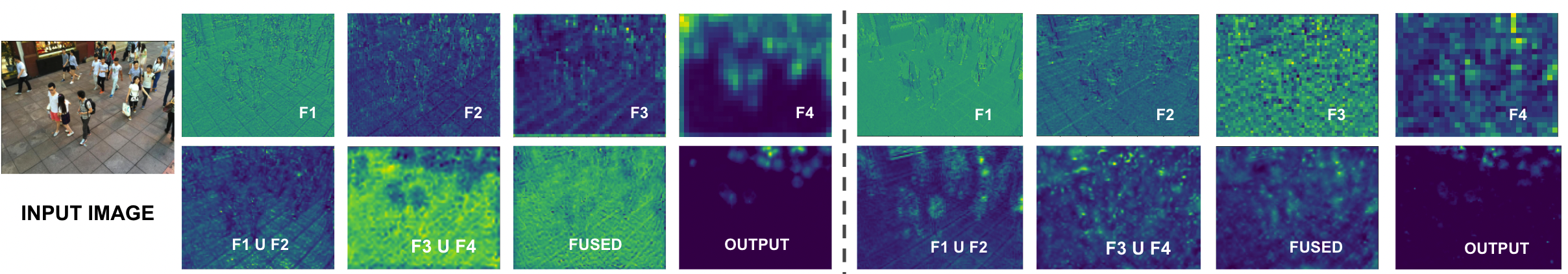}
  \captionsetup{font=small}
  \caption{\small Left: Feature maps from each layer of ASFNet-B,  Right: Feature maps from each layer of ASFNet-S. Where F1, F2, F3 and F4 shows the different scale features and features from intermediate layers.}
  \label{figure_2}
\end{figure*}
\subsection{Adjacent Featutres}
Given an input image $I_{i} \in \mathbb{R}^{C_i \times H_i \times W_i}$, where $C_i$ refers to the number of channels, $H_i$ is Height, and $W_i$ is the width of the image, $I_i$ is fed into the backbone network to obtain four different sets of multi-scale feature maps \{$F_1, F_2, F_3, F_4$\} from intermediate layers as shown in Figure \ref{figure_2}. Using larger kernel sizes in CNN increases the number of parameters and  FLOPs. This happens because larger kernels have more weights to learn, resulting in a higher parameter count, and they require more computations during forward and backward passes, leading to increased FLOPs. Therefore, it's important to strike a balance between kernel size and computational efficiency when designing CNN architectures to avoid unnecessary complexity and resource usage, so we set  \{(3x3, dil = 2), (3x3), (1x1), (1x1)\} as kernels for the given four sets of feature maps. After applying the convolution operation $\mathcal{H}_i$ followed by ReLU activation $\gamma$, we multiply each feature $F^i$ by a parameter $\lambda_i$, which reduces the noise in the feature map. 
\begin{equation}
F_{i}^{'}  = \lambda_i(\gamma\left(\mathcal{H}_{i}(F_i)\right))
\end{equation}
where  i,k={1,2,3,4}.
We then apply bicubic interpolation for every feature map to rescale every feature map to a similar dimension. The interpolated pixel value  at coordinates \((x, y)\) can be expressed using bicubic interpolation as follows:
\begin{equation}
 \mathcal{F}_{inter}(x, y) = \sum_{i=0}^{3} \sum_{j=0}^{3} w_{ij} \cdot F_{k}(x + i, y + j)
\end{equation}
Where $F_k(x + i, y + j)$ represents the pixel values of the neighbouring pixels within the local $4 \times 4$ neighbourhood around $(x, y)$ and $w_{ij}$ are the bicubic interpolation weights. Then we concatenate adjacent scaled features map $F_{1}^{'}$,$F_{2}^{'}$ and $F_{3}^{'}$,$F_{4}^{'}$ to get \{ $f_{fuse}^{1}$, $f_{fuse}^{2}$\} as a set of new feature maps; We do this to process similar complexity of semantics at each level. The process continues till we reach the final set of feature maps $f_{net}$ that goes through 1x1 convolutions to generate the final density map. The mathematics is as described below.

\begin{equation}
    f_{fuse}^{i}(a,b)= \gamma(\mathcal{H}_{i}(F_{a}^{'}  \cup F_{b}^{'} ))
\end{equation}
\begin{equation}
    F_{fused}= \gamma(\mathcal{H}_{fused}(\bigcup_{i=1}^{2} f_{fuse}^{i}(2i,2i-1)))
\end{equation}
\begin{equation}
    f_{net}= \gamma(\mathcal{H}_{net}(F_{fused}))
\end{equation}
\subsection{Loss Function}
Our final output is a 2-dimensional density map; hence, we use a pixel-wise L2 loss function that compares each pixel value in the density map corresponding to its ground truth.
\begin{equation}
L(\Theta) = \frac{1}{2n} \sum_{i=1}^{n} \left\| h(\theta, x_i) - G \right\|_{2}^2
\end{equation}
Where n is the size of the training set, and $x_i$ represents the $i_{th}$ input image. $\text{h}(\Theta,x_i)$ is the output and $\text{G}$ is ground truth density map of image. $L(\Theta)$  denotes the loss between the ground-truth density map and the estimated density map.

\begin{table*}[ht!]
\label{tab_1}
\resizebox{\textwidth}{!}{%
\begin{tabular}{|l|ll|ll|ll|l|l|l|l|}
\hline
\multirow{2}{*}{Methods} &
  \multicolumn{2}{l|}{ShanghaiTech A} &
  \multicolumn{2}{l|}{ShanghaiTech B} &
  \multicolumn{2}{l|}{UCF-CC-50} &
  \multirow{2}{*}{\#Param} &
  \multirow{2}{*}{FLOPs} &
  \multirow{2}{*}{Size(MB)} &
  \multirow{2}{*}{Infrence(s)\*} \\ \cline{2-7}
         & \multicolumn{1}{l|}{MSE}   & MAE   & \multicolumn{1}{l|}{MSE}   & MAE   & \multicolumn{1}{l|}{MSE}   & MAE    &       &         &      &   \\ \hline
CSRNet\cite{li2018csrnet}   & \multicolumn{1}{l|}{115.0} & 68.2  & \multicolumn{1}{l|}{16.0}  & 10.6  & \multicolumn{1}{l|}{397.5} & 266.1  & 16.26 & 649.813 & 62.05    & 36.56 \\
CANNet\cite{liu2019context}   & \multicolumn{1}{l|}{100}   & 62.3  & \multicolumn{1}{l|}{12.2}  & 7.8   & \multicolumn{1}{l|}{243.7} & 212.2  & 18.12 & 688.526  & 69.09    & 90.34 \\
ASPDNet\cite{gao2020counting}  & \multicolumn{1}{l|}{96.2}  & 60.8  & \multicolumn{1}{l|}{10.5}  & 7.2   & \multicolumn{1}{l|}{-}     & -      & 27.42 & 911.108   & 104.63    & 255 \\
SCARNet\cite{gao2019scar}  & \multicolumn{1}{l|}{144.1} & 66.3  & \multicolumn{1}{l|}{15.2}  & 9.5   & \multicolumn{1}{l|}{-}     & -      & 16.28 & 650.388  & 62.16    & 84.98\\

P2PNet\cite{song2021rethinking} & \multicolumn{1}{l|}{\textcolor{blue}{85.1}} &\textcolor{blue}{52.7} & \multicolumn{1}{l|}{\textcolor{blue}{9.9}}    & \textcolor{blue}{6.3}  & \multicolumn{1}{l|}{256.1} & 172.7  & 18.34 & -   & 82.3 & - \\ 
UEPNet\cite{wang2021uniformity}  & \multicolumn{1}{l|}{91.2}   & 54.6  & \multicolumn{1}{l|}{10.9}    & 6.4    & \multicolumn{1}{l|}{\textcolor{blue}{131.7}} & \textcolor{blue}{81.1}  & 26.21  &         & -    & - \\ \hline
SANet\cite{cao2018scale}   & \multicolumn{1}{l|}{122.2} & 75.3 & \multicolumn{1}{l|}{17.9}   & 10.5  & \multicolumn{1}{l|}{334.9} & 258.4 & 0.91 & -  & -    & - \\

MCNN\cite{zhang2016single}     & \multicolumn{1}{l|}{173.2} & 110.2 & \multicolumn{1}{l|}{41.3}    & 26.4  & \multicolumn{1}{l|}{509.1} & 377.6  & 0.133 & 42.26    & 0.53 & 5.38 \\ 
TDFCNN\cite{sam2018top}   & \multicolumn{1}{l|}{145.1}   & 97.5  & \multicolumn{1}{l|}{32.8}    & 20.7    & \multicolumn{1}{l|}{491.4} & 354.7  &0.13  &    -    & -    & - \\ 

C-CNN\cite{shi2020real}   & \multicolumn{1}{l|}{141.7}   & 88.1  & \multicolumn{1}{l|}{22.1}    & 14.9    & \multicolumn{1}{l|}{-} & -  & \textcolor{blue}{0.073}  &   19.830   &\textcolor{blue}{0.23}   & \textcolor{blue}{0.8} \\ 

ACSCP\cite{Shen_2018_CVPR} & \multicolumn{1}{l|}{102.7} & 75.7 & \multicolumn{1}{l|}{27.4}    & 17.2  & \multicolumn{1}{l|}{404.6} & 291.0 & {5.10} & -   & - & - \\ 

\textbf{ASFNet-S} & \multicolumn{1}{l|}{91.67} & 61.09 & \multicolumn{1}{l|}{18}    & 11.2  & \multicolumn{1}{l|}{252.3} & 192.1  & {2.58} & \textcolor{red}{2.185}   & 15.5 & 27 \\ 

\textbf{ASFNet-B} & \multicolumn{1}{l|} {\textcolor{red}{88.67}} & \textcolor{red}{59.32} & \multicolumn{1}{l|}{\textcolor{red}{11.02}} & \textcolor{red}{8.2}  & \multicolumn{1}{l|}{\textcolor{red}{151.49}}     & \textcolor{red}{106.69}    & 5.69  &  29.803   & 10.6 & 71.2 \\ \hline
\end{tabular}%
}
\captionsetup{font=small}
  \caption{\small Comparison with state-of-the-art methods. \#Param denotes the number of parameters, while FLOPs is the number of Floating Point Operations, and Size is the size of the trained model. Execution time is computed on an Nvidia T400 GPU. The units are million (M) for \#Param, giga (G) for FLOPs, millisecond (ms) for GPU time, and size is in MegaByte (MB). The entire table is divided into two parts: the upper one contains heavy models, while the lower part contains lightweight models.
  Blue colour shows the SOTA result, and red shows the best performance among lightweight models. }
\end{table*}
\vspace{-3.0mm}
\section{Experiments}
\vspace{-1mm}
\textbf{Dataset:}Our experiments are conducted on three benchmark datasets: ShanghaiTech (Part A and Part B) and the UCF-CC-50 dataset. In ShanghaiTech Part A, a total of 482 crowd images were divided into 300 training images and 182 testing images. The average number of pedestrians in these images was approximately 501. Meanwhile, ShanghaiTech Part B contained 716 images, with 400 allocated for training and 316 for testing. In contrast to Part A, the average pedestrian count in Part B was notably smaller, averaging around 123 pedestrians per image. The UCF CC 50 dataset consisted of 50 images with high crowd density, featuring varying pedestrian counts, ranging from 94 to 4,543 pedestrians per image.\\
\textbf{Ground Truth Generation:-} We follow\cite{zhang2016single} To generate a ground truth density map that uses the geometry-adaptive kernels; due to invariance in the crowd,  we generate the ground truth density maps with the spatial distribution information across the whole image. The geometric adaptive Gaussian kernel is given as
\vspace{-0.8mm}
\begin{equation}
F(x) = \sum_{i=1}^{N} \delta(x - x_i) \ast G_{\sigma_i}, \quad \sigma_i = \beta \bar{d_i} 
\end{equation}
where $\sigma_i = \beta \bar{d_i}$,  $x_i$ represents a specific target object within the ground truth $\delta$. To transform $\delta(x - x_i)$ into a density map, we convolve it with a Gaussian kernel having a standard deviation of $\sigma_i$. Here, $\bar{d_i}$ signifies the average distance among the $k$ nearest neighbours of the target object $x_i$; for our experiment, we set k = $10$.\\
\textbf{Evalution metrics:-} Followed by the previous crowd counting model \cite{zhang2016single} \cite{liu2019context}, we use Mean Squared Error(MSE) and Mean absolute error(MAE) standard measures for evaluating the effectiveness and accuracy of our model. To assess the lightweightness,  number of parameters, Floating Point Operation (FLOP) and model size are considered as well-suited metrics.\\
\textbf{Training Details:} 
We train both models using the same hyperparameters and as the Adam optimizer with a learning rate of $5e-5$ and weight decay as $1e-4$. After experimenting with various combinations, our experiments have empirically shown that appropriate values of $\lambda_1$=0.1, $\lambda_2$=0.1, $\lambda_3$=0.5, $\lambda_4$=1.0
We train both models on Tesla T4 GPU for 500 epochs.
\vspace{-5mm}
\section{ Results}
\vspace{-2mm}
We evaluate our proposed lightweight models with previous state-of-the-art (SOTA) work on three datasets, as shown in Table 1 , considering key metrics such as MSE, MAE, inference time, number of parameters, and model size. Our findings show that the proposed models acquire a balance between performance and efficiency. ASFNet-B achieves comparable results to the SOTA across all three datasets while utilizing fewer parameters, requiring lower FLOPs, exhibiting a smaller model size, and demonstrating faster inference times. Meanwhile, ASFNet-S excels in the lightweight category, particularly in terms of MSE, MAE, and FLOPs, however, with a slight increase in parameters and model size. To the best of our knowledge, we have the lowest FLOPs in comparison to existing works.
 \vspace{-0.25cm}
\section{Ablation Study}
To evaluate the effectiveness of our approaches, we conducted a series of experiments for ablation study, as shown in Table\ref{table_2}, where we fused different feature groups instead of adjacent features and found that fusing adjacent features gives the best results. We conducted experiments to assess the impact of weighted convolutions by comparing the model's accuracy when trained with and without the use of weight and it has been observed a slight decrease in the model's performance when we excluded the convolutional weights from the training process. Additionally, we examined the effects of 25\% L1 and 25\% L2 pruning techniques on the model's performance as shown in Table \ref{table_3}

\begin{table}[htbp]
\centering
\caption{\small Ablation Study of ASFNet-B on ShanghaiTech-B Benchmark Dataset with different feature set fusions and non-weighted convolutions.}
\label{table_2}
\resizebox{0.32\textwidth}{!}{%
\begin{tabular}{|l|ll|}
\hline
\multirow{2}{*}{\begin{tabular}[c]{@{}l@{}}Combination(ASFNet-B)\end{tabular}} & \multicolumn{2}{l|}{ShanghaiTech-B} \\ \cline{2-3} 
                                                                                           & \multicolumn{1}{l|}{MSE}         & MAE       \\ \hline
fuse(F1, F3), fuse(F2, F4)                                                                 & \multicolumn{1}{l|}{14.07}       & 8.61      \\ \hline
fuse(F1, F4), fuse(F2, F3)                                                                 & \multicolumn{1}{l|}{13.32}       & 8.48      \\ \hline
fuse(F1, F2), fuse(F3, F4)                                                                 & \multicolumn{1}{l|}{11.02}       & 8.2       \\ \hline
No weights                                                                                 & \multicolumn{1}{l|}{13.75}       & 8.12      \\ \hline
\end{tabular}%
}

\end{table}
\vspace{-6mm}

\begin{table}[h!]
\centering
\captionsetup{font=small}
  \caption{\small L1 Pruning and L2 Pruning on ASF-Net-B. Evaluated on ShanghaiTech-B Dataset}
\label{table_3}
\resizebox{0.32\textwidth}{!}{%
\begin{tabular}{|l|ll|ll|}
\hline
\multirow{2}{*}{DATASET} & \multicolumn{2}{l|}{L1 Pruning} & \multicolumn{2}{l|}{L2 Pruning} \\ \cline{2-5} 
             & \multicolumn{1}{l|}{MSE}   & MAE   & \multicolumn{1}{l|}{MAE}   & MSE    \\ \hline
ShanghaiTech A & \multicolumn{1}{l|}{118.1} & 73.2  & \multicolumn{1}{l|}{160.1} & 92.8   \\
ShanghaiTech B  & \multicolumn{1}{l|}{32.25} & 17.52 & \multicolumn{1}{l|}{34.57}      &   19.60     \\
UCF-CC-50    & \multicolumn{1}{l|}{390}   & 270.2 & \multicolumn{1}{l|}{440.7} & 312.21 \\ \hline
\end{tabular}%
}

\end{table}
\vspace{-6mm}
\section{Conclusion}
\vspace{-2mm}
In this work, we propose a simple yet computationally efficient model called ASFNet, giving comparable accuracy to multiple SOTA models at substantially lesser parameter count, FLOPs count, and model size. We support our claims with extensive experiments. 
Although The models prove efficient, the ASFNet-B model has a slightly larger inference time than its competitors. Reducing inference time in ViT-based crowd regression architectures can be a future research direction.


\bibliographystyle{IEEEbib}

{\footnotesize 
\bibliography{strings,refrer}
}

\end{document}